%% file: main.tex
\documentclass[conference]{IEEEtran}

\IEEEoverridecommandlockouts

\usepackage{textcomp}
\usepackage[numbers]{natbib}
\usepackage[most,skins,theorems]{tcolorbox}

\tcbset{
  aibox/.style={
    width=\linewidth,
    top=8pt,
    bottom=4pt,
    colback=blue!6!white,
    colframe=black,
    colbacktitle=black,
    enhanced,
    center,
    attach boxed title to top left={yshift=-0.1in,xshift=0.15in},
    boxed title style={boxrule=0pt,colframe=white,},
  }
}
\newtcolorbox{AIbox}[2][]{aibox,title=#2,#1}
\definecolor{lightblue}{rgb}{0.22,0.45,0.70}%

\usepackage{xcolor}
\definecolor{gblue}{HTML}{4f79a7}
\definecolor{ggreen}{HTML}{77b7b2}
\definecolor{gred}{HTML}{e1575a}
\definecolor{gorange}{HTML}{f28e2a}
\definecolor{ppurple}{HTML}{603a70}
\definecolor{mydarkblue}{rgb}{0,0.08,0.45}
\usepackage[colorlinks=true,
            citecolor=mydarkblue,
            linkcolor=blue,
            anchorcolor=red,
            urlcolor=blue
            ]{hyperref}

\usepackage[utf8]{inputenc} %
\usepackage[T1]{fontenc}    %
\usepackage{url}            %
\usepackage{booktabs}       %
\usepackage{amsfonts}       %
\usepackage{nicefrac}       %
\usepackage{microtype}      %
\usepackage{lipsum}		%
\usepackage{graphicx}
\usepackage{doi}

\usepackage[utf8]{inputenc} %
\usepackage[T1]{fontenc}    %
\usepackage{url}            %
\usepackage{booktabs}       %
\usepackage{amsfonts}       %
\usepackage{nicefrac}       %
\usepackage{microtype}      %

\usepackage{multirow}
\usepackage{graphicx}
\usepackage[font=small,labelfont=bf]{caption}
\usepackage{subcaption}
\usepackage{amsmath, amsthm, amssymb}

\usepackage[subtle]{savetrees}

\usepackage{mathtools}
\usepackage{xspace}
\usepackage{enumitem}
\usepackage{footnote}
\usepackage{color}

\usepackage{thmtools}
\usepackage{thm-restate}

\usepackage{csquotes}
\usepackage{wrapfig}

\usepackage{multicol}
\usepackage{lipsum} %
\usepackage{tabularx}

\usepackage{algorithm, algorithmic}
\usepackage[algo2e,ruled,vlined]{algorithm2e}

\usepackage{tablefootnote}

\usepackage{peanutgallery}

\usepackage{color-edits}
\addauthor{sw}{blue}

\usepackage[most,skins,theorems]{tcolorbox}

\tcbset{
  aibox/.style={
    width=\linewidth,
    top=8pt,
    bottom=4pt,
    colback=ppurple!6!white,
    colframe=ppurple,
    colbacktitle=ppurple,
    enhanced,
    center,
    attach boxed title to top left={yshift=-0.1in,xshift=0.15in},
    boxed title style={boxrule=0pt,colframe=white,},
  }
}

\def\BibTeX{{\rm B\kern-.05em{\sc i\kern-.025em b}\kern-.08em
    T\kern-.1667em\lower.7ex\hbox{E}\kern-.125emX}}
\begin{document}

\title{Position: LLM Unlearning Benchmarks\\ are Weak Measures of Progress 
}

\author{\IEEEauthorblockN{Pratiksha Thaker, Shengyuan Hu, Neil Kale, Yash Maurya, Zhiwei Steven Wu, Virginia Smith}
\textit{Carnegie Mellon University}\\
Pittsburgh, PA \\
\small\texttt{\{pthaker, shengyua, nkale, ymaurya, zstevenwu, smithv\}@andrew.cmu.edu}}

\maketitle

\begin{abstract}
\input{tex-src/abstract}
\end{abstract}

\begin{IEEEkeywords}
machine unlearning, LLMs, benchmarking
\end{IEEEkeywords}

\input{tex-src/intro}

\input{tex-src/related}
\input{tex-src/threatmodel}
\input{tex-src/adversarial-benchmarks}
\input{tex-src/training-on-test}
\input{tex-src/recommendations}
\input{tex-src/conclusion}

\bibliographystyle{IEEEtranN}
\bibliography{unlearn}

\appendix
\input{tex-src/appendix-wmdp}

\end{document}

%% file: tex-src/abstract.tex
Unlearning methods have the potential to improve the privacy and safety of large language models (LLMs) by removing sensitive or harmful information post hoc.
The LLM unlearning research community has increasingly turned toward empirical benchmarks to assess the effectiveness of such methods.
In this paper, we find that existing benchmarks provide an overly optimistic and potentially misleading view on the effectiveness of candidate unlearning methods.
By introducing simple, benign modifications to a number of popular benchmarks, we expose instances where supposedly unlearned information remains accessible, or where the unlearning process has degraded the model’s performance on retained information to a much greater extent than indicated by the original benchmark.
We identify that existing benchmarks are particularly vulnerable to modifications that introduce even loose dependencies between the forget and retain information.
Further, we show that ambiguity in unlearning targets in existing benchmarks can easily lead to the design of methods that overfit to the given test queries. Based on our findings, we urge the community to be cautious when interpreting benchmark results as reliable measures of progress, and we provide several recommendations to guide future LLM unlearning research.

%% file: tex-src/intro.tex
\vspace{.1in}
\section{Introduction}
\label{sec:intro}

Large-scale data collection, particularly through data available on the Web,
has enabled stunning progress in the capabilities of generative models over the past decade,
but also highlighted the potential cost to user privacy in the process.
In light of increasing concern about the indiscriminate use of potentially private data to train generative models, 
particularly large language models,
a growing body of work has explored approaches for data ``unlearning,'' which aim to remove the influence of specific subsets of training data
\cite{bourtoule2021machine, gupta2021adaptive, ginart2019making, muresanu2024unlearnable}
or prevent generations about specific topics \cite{li2024wmdp, jin2024rwku, eldan2023s}. %

Unlearning in LLMs is particularly challenging given that
the models and training datasets (if available) are massive, and the information to be unlearned may not be contained to specific data points~\cite{brown2022does, tramer2022considerations}. Many research efforts have thus focused on the development of methods for \textit{approximate unlearning}, which aim to efficiently approximate the effect of retraining a model from scratch on a dataset with all offending training data removed.
Evaluating the success of an approximate unlearning algorithm on an LLM can be difficult
due to the scale of both the data and the model,
leading researchers to instead develop benchmarks that rely on carefully curated test queries used to probe the knowledge of the unlearned model. %
These benchmarks commonly consist of:
\begin{itemize}
\item A \textit{forget set} of test queries intended to measure whether specific data or knowledge has been unlearned.
\item A \textit{retain set} of test queries intended to ensure retention of data unrelated to the unlearning data. %
\end{itemize}
\vspace{-.05in}

For example, benchmarks have been developed that aim to use forget and retain queries to assess whether an unlearning method has effectively forgotten text corresponding to the book series \emph{Harry Potter}~\cite{eldan2023s}; information about specific individuals in a synthetic dataset~\cite{maini2024tofu}; or knowledge that could give information to malicious agents~\cite{li2024wmdp}, all while retaining more general knowledge in the LLM.

The field of unlearning (particularly in LLMs) is relatively new, 
and the fact that several benchmarks already exist is remarkable.
In the absence of provable guarantees, 
empirical benchmarks would seem to be the best measure of progress in this area.
Indeed, we find that many unlearning papers claim success exclusively based on performance on a few benchmarks (see Section~\ref{sec:relatedwork}). %
As these benchmarks are increasingly influential in the community,
it is important to examine them and ensure that benchmark performance correlates accurately with the stated goals of unlearning.
Even without an agreed-upon formal definition of unlearning in LLMs, 
we should at least hope that benchmark performance matches our high-level intuitions about what unlearning should achieve.

Unfortunately, we find through a wide range of experiments 
 on popular LLM unlearning benchmarks that 
even small, non-adversarial modifications to benchmark data
can uncover model behavior that contradicts the reported claims of successful unlearning.
As a result, we caution the community about generalizing an algorithm's unlearning performance
beyond the very specific evaluation queries themselves. Our results here echo recent works scrutinizing other, more general LLM evaluation benchmarks, in that it is difficult to produce a completely exhaustive set of queries that measure the model's knowledge of a certain subject~\cite[e.g.,][]{zheng2023large, alzahrani2024benchmarks, wang2023large, zhao2021calibrate, pezeshkpour2023large, debenedetti2024dataset}. However, we also highlight two key concerns specific to LLM unlearning evaluation:

First, we identify that unlearning algorithms are particularly vulnerable in the face of queries that have dependencies on both the unlearned data and retained data.
Popular existing benchmarks typically evaluate on  \emph{disjoint} forget and retain sets (see e.g. ~\cite{liu2024rethinking}, Table 1), 
where forget queries are related only to the unlearned data
and retain queries are largely unrelated to the unlearned data. %
As a result, we find that simple modifications that introduce even rough dependencies between the forget and retain sets can cause catastrophic failures in unlearning methods optimized on these benchmarks.

Second, we find that existing LLM unlearning benchmarks implicitly encourage training on or fitting to the evaluation query set, for two reasons.
First, effective solutions may involve pre- or post-processing the queries themselves,
but in the absence of a held-out query set, developing such a solution involves inspecting the test queries.
Furthermore, in some settings where the goal is to unlearn a general concept rather than specific training data, 
no well-defined set of data to unlearn is even provided,
and the information to unlearn is instead implicitly \emph{defined by the test set} (the forget and retain queries).
In other cases, the data to unlearn is provided,
but it is unclear what effect unlearning should have on test metrics; for example, if some test retain set questions are correlated with the unlearn data,
then effective unlearning should \emph{reduce} accuracy on the retain set. We show these ambiguities have resulted in the development of a number of unlearning approaches which may seem to excel on benchmark performance alone, but are woefully ill-equipped to handle LLM unlearning in practical scenarios.

Overall, our analysis across a wide range of popular LLM unlearning benchmarks leads us to the following key position:

\vspace{.1in}
\textit{\textbf{Main position:} Existing empirical benchmarks for LLM unlearning are limited measures of progress in the best case, and are actively misleading in the worst case. We recommend that the LLM unlearning community thus proceed with caution when interpreting the results of these benchmarks alone, and should focus on crafting definitions and metrics that more carefully match practical use cases, as well as provable algorithms that meet these definitions.}
\smallskip

We note that although we describe attacks on specific unlearning algorithms in our paper, 
our intent is not to discredit or pick on any specific method,
but rather to provide examples of how general properties of unlearning benchmarks can lead to counterintuitive or unexpected results. Similarly, we do not aim to discredit any one LLM unlearning benchmark, but rather show that many benchmarks share common properties that can result in misleading advances in unlearning (as we discuss in Section~\ref{sec:recommendations}, developing fully comprehensive LLM unlearning benchmarks is an extremely challenging task). %

%% file: tex-src/related.tex
\section{Motivation: The Rise of LLM Unlearning Benchmarks}
\label{sec:relatedwork}
We first motivate our study by exploring recent trends in approximate LLM unlearning evaluation. We note that our paper is not intended to be a survey paper and as such we do not claim to have exhaustively covered the space of unlearning work published particularly in the past two years. 
We point the reader to a number of comprehensive surveys on unlearning 
for further background, including~\cite{thudi2022unrolling, xu2023machine, hayes2024inexact, zhang2024right, si2023knowledge, qu2024frontier, xu2024machine, blanco2024digital, liu2024machine, shintre2019making}.

\begin{figure*}[t!]
    \centering
    \begin{subfigure}[t]{0.48\textwidth}
        \centering
        \includegraphics[clip,width=\textwidth]{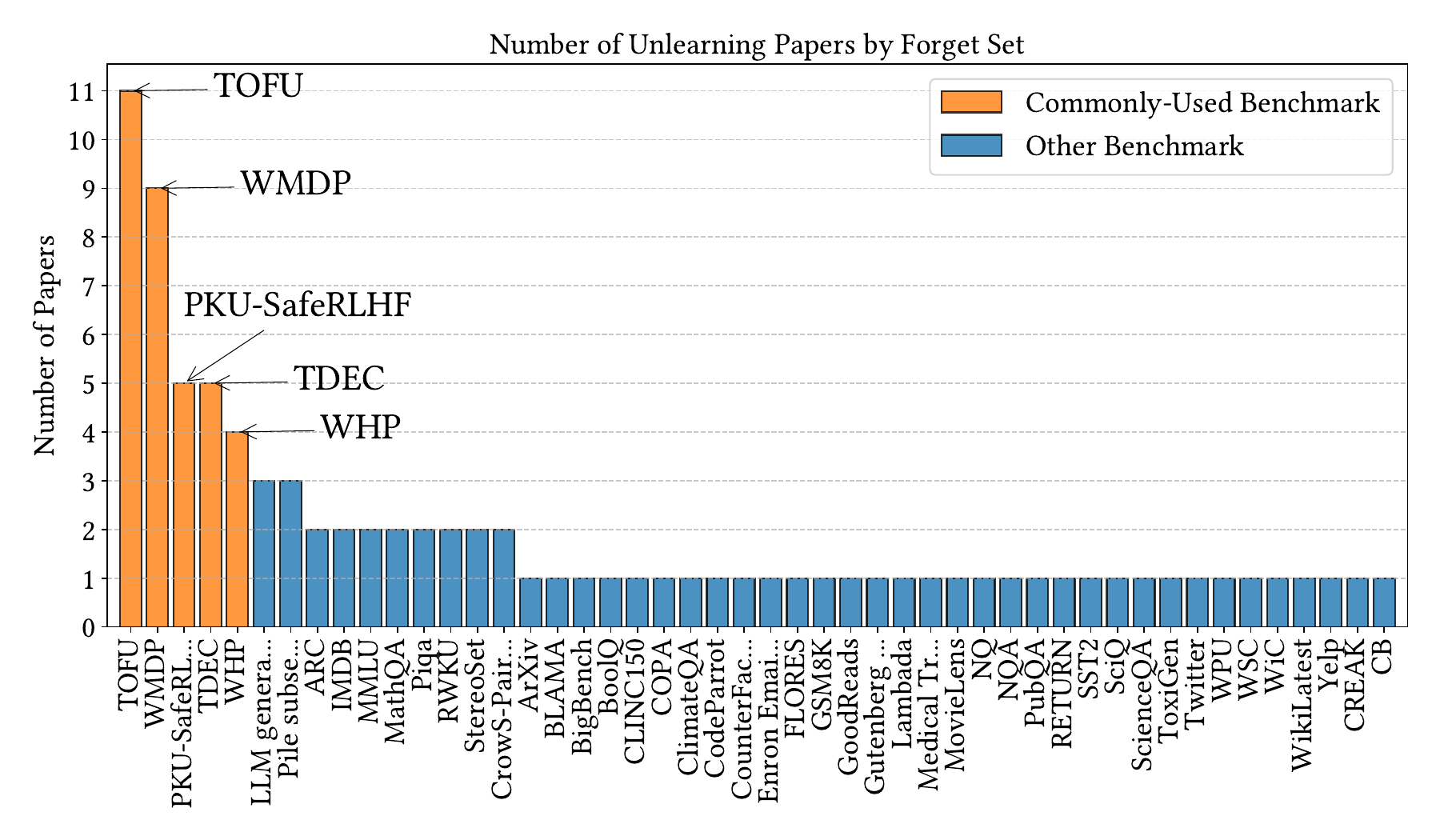}
        \caption{Number of unlearning papers by forget set}
        \label{fig:forget-set-hist}
    \end{subfigure}\hfill
    \begin{subfigure}[t]{0.48\textwidth}
        \centering
        \includegraphics[clip,width=\textwidth]{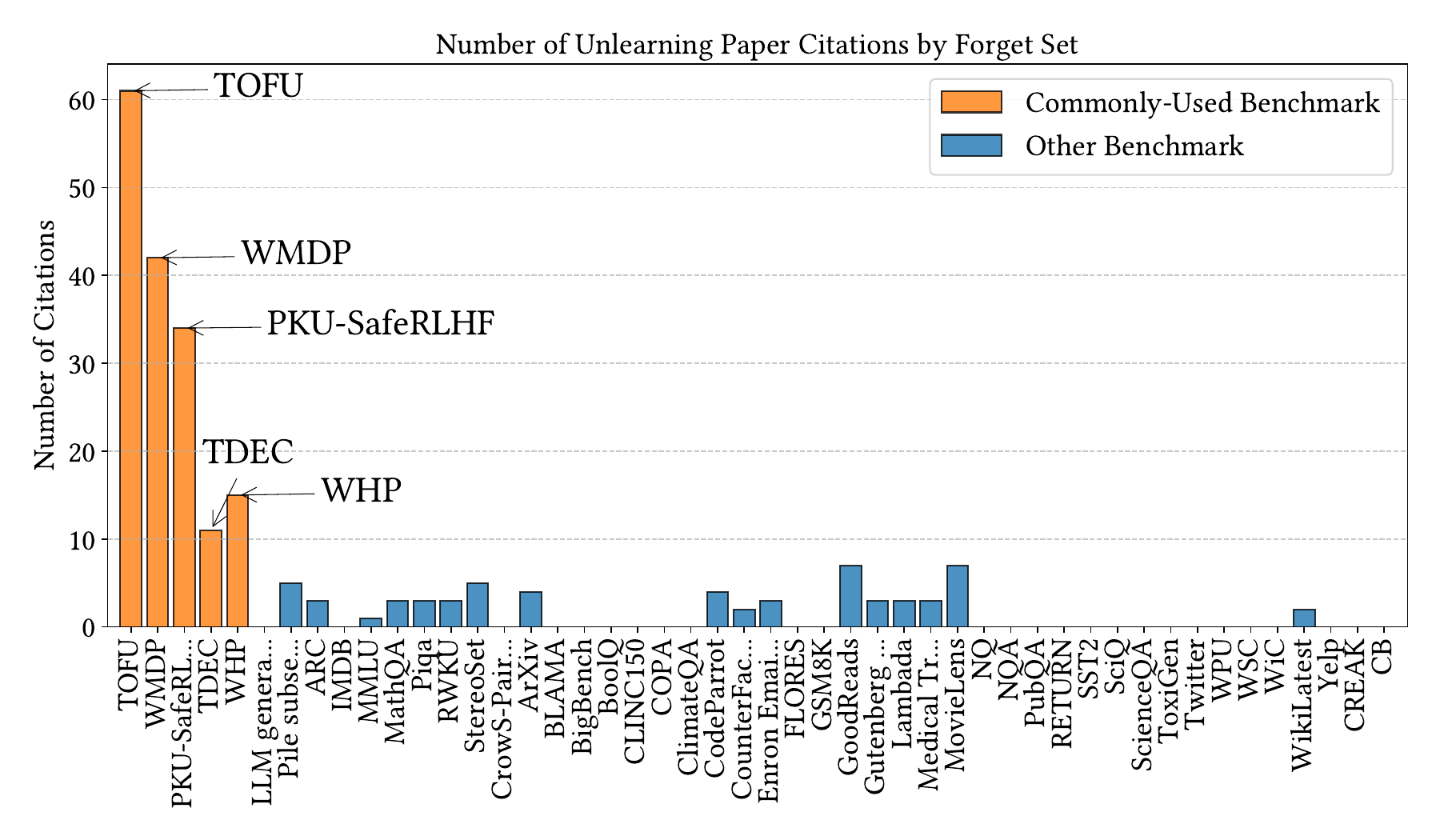}
        \caption{Number of unlearning paper citations by forget set}
        \label{fig:forget-set-citations}
    \end{subfigure}
    \caption{We survey works on LLM unlearning published in 2024 and find that they predominantly evaluate unlearning success on fixed ``forget set'' benchmarks~\cite{liu2024awesome}. \textit{\textbf{(a)}} Notably, the 5 most commonly-used benchmarks (red) account for nearly half of all evaluations, with TOFU \cite{maini2024tofu} and WMDP \cite{li2024wmdp} alone used in 31\% of papers. \textit{\textbf{(b)}} Additionally, papers evaluating on these common benchmarks (red) receive over 80\% of total citations across the repository. %
    }
    \label{fig:forget-sets}
\end{figure*}

\subsection{Focus of this work: forget set/retain set benchmarks}

To evaluate machine unlearning, the `gold standard' approach involves comparing the unlearned model to a model trained from scratch with all target unlearning data removed. Unfortunately, such comparisons are typically infeasible for LLMs, as the training datasets are frequently inaccessible, and, even if they were available, retraining would incur a significant computational expense. 
To address this challenge, an increasingly popular approach has been to develop benchmarks that instead evaluate the outcomes of unlearning by prompting the LLM with a specified set of ``forget set'' and ``retain set'' queries.

We note that forget/retain set evaluation does not cover the entire space of unlearning evaluation:
for example, 
some work focuses more narrowly on data memorization,
showing that data cannot be recovered exactly by prompting with partial prefixes~\cite{huang2024demystifying, nasr2023scalable}, that memorized data can be localized within model weights~\cite{stoehr2024localizing, patil2023can, chang2023localization, hase2023does}, or that certain tokens in the sequence have very low probability after unlearning ~\cite{ashuach2024revs}. 
However, inspired by formal definitions of unlearning and algorithms for unlearning in simpler models \cite{neel2021descent},
many works target the more ambitious goal of entirely removing the influence of some training data from the model. Here, we find that the forget set/retain set structure is increasingly popular and influential in the literature.

Indeed, to quantify this influence and justify our choice of focus, we survey the 72 unlearning papers from 2024 included in the open-source Awesome LLM Unlearning repository~\cite{liu2024awesome}.
While we recognize that this repository could reflect biases inherent to its curation process or prevailing trends in the field, the results provide a valuable indicator of broader research patterns.
Within this set, an overwhelming fraction (82\%) of papers have an evaluation following the forget set/retain set structure.

In addition, we find that a small handful of forget/retain benchmarks seem to have an outsized influence on evaluations; we thus focus on these benchmarks in our work. In Figure~\ref{fig:forget-set-hist}, we plot the prevalence of each forget set we find in the literature.
It is clear that a handful of benchmarks are overwhelmingly popular:
TOFU~\cite{maini2024tofu}, WMDP~\cite{li2024wmdp}, TDEC~\cite{google2023tdec}, PKU-SafeRLHF~\cite{ji2024pku}, and Who's Harry Potter~\cite{eldan2023s}. Almost half of the papers in the repository use one of these five benchmarks to evaluate unlearning performance.

To further analyze the impact of these particular benchmarks,
we additionally consider the count of citations of the papers in our survey set in Figure~\ref{fig:forget-set-citations}.
Here we find that papers evaluating on the top 5 forget sets account for 80\% of the total citations attributable to papers in the survey set, emphasizing the influence of these benchmarks on unlearning evaluation and the direction of the community at large.

\subsection{Description of common unlearning benchmarks}

Motivated by the analysis above, we briefly describe the top five most popular LLM unlearning benchmarks, which all evaluate on ``forget set'' and ``retain set'' queries. %

\textbf{``Who's Harry Potter?''~\cite{eldan2023s}.} \citet{eldan2023s} introduced a benchmark in 2023 for unlearning information from the \emph{Harry Potter} book series.
To evaluate unlearning, the authors query a set of 300 \emph{Harry Potter}-related questions
and evaluate them automatically using a score from 0 to 5 computed using a GPT-4 evaluator.
To evaluate model quality on non-\emph{Harry Potter}-related information,
the authors run a number of standard benchmarks such as HellaSwag~\cite{zellers2019hellaswag} and OpenBookQA~\cite{mihaylov2018can}.

\textbf{TOFU~\cite{maini2024tofu}.} The more recent TOFU benchmark ~\cite{maini2024tofu}
consists of a synthetic dataset of 4000 questions and answers about fictional authors.
The authors provide a version of llama-2-7b-chat-hf that is finetuned on the synthetic dataset, 
and the goal is to unlearn subsets (1\%, 5\%, or 10\%) of the authors' information from the finetuned model.

\textbf{WMDP~\cite{li2024wmdp}.} The Weapons of Mass Destruction Proxy (WMDP)
benchmark~\cite{li2024wmdp} aims to simulate forgetting ``expert-level'' knowledge that could be used to create bioweapons or launch cyberattacks.
The forget set consists of multiple-choice expert-level questions about biology, cybersecurity, and chemistry,
and the retain set is drawn from MMLU~\cite{hendrycks2020measuring} college biology, virology, college computer science, and computer security question sets.
The primary metrics considered are accuracy on these question sets.

\textbf{RWKU~\cite{jin2024rwku}.} The Real World Knowledge Unlearning (RWKU) benchmark \cite{jin2024rwku} contains a dataset of 13131 synthetic questions relating to real-world celebrities. The goal is to unlearn information about 200 celebrities from the original model. The forget benchmark includes a variety of adversarial probes such as knowledge manipulation, knowledge memorization, and membership inference attacks. The retain benchmark measures model utility on five capabilities such as reasoning ability measured by Big-Bench-Hard \cite{suzgun2022challenging} and truthfulness measured on TruthfulQA \cite{lin2021truthfulqa}.

\textbf{TDEC~\cite{google2023tdec}.} Finally, the Training Data Extraction Challenge (TDEC) is a subset of the Pile dataset \cite{gao2020pile} containing 20,000 easy-to-extract examples. The TDEC benchmark is primarily used to evaluate unlearning defenses against exact sequence extraction attacks \cite{wang2024selective, barbulescu2024each, cha2024towards}. Although TDEC is among the more popular benchmarks, 
our focus in this work is on broader evaluations compared to data extraction/memorization alone.

\begin{figure*}[t!]
    \centering
    \includegraphics[width=0.85\textwidth]{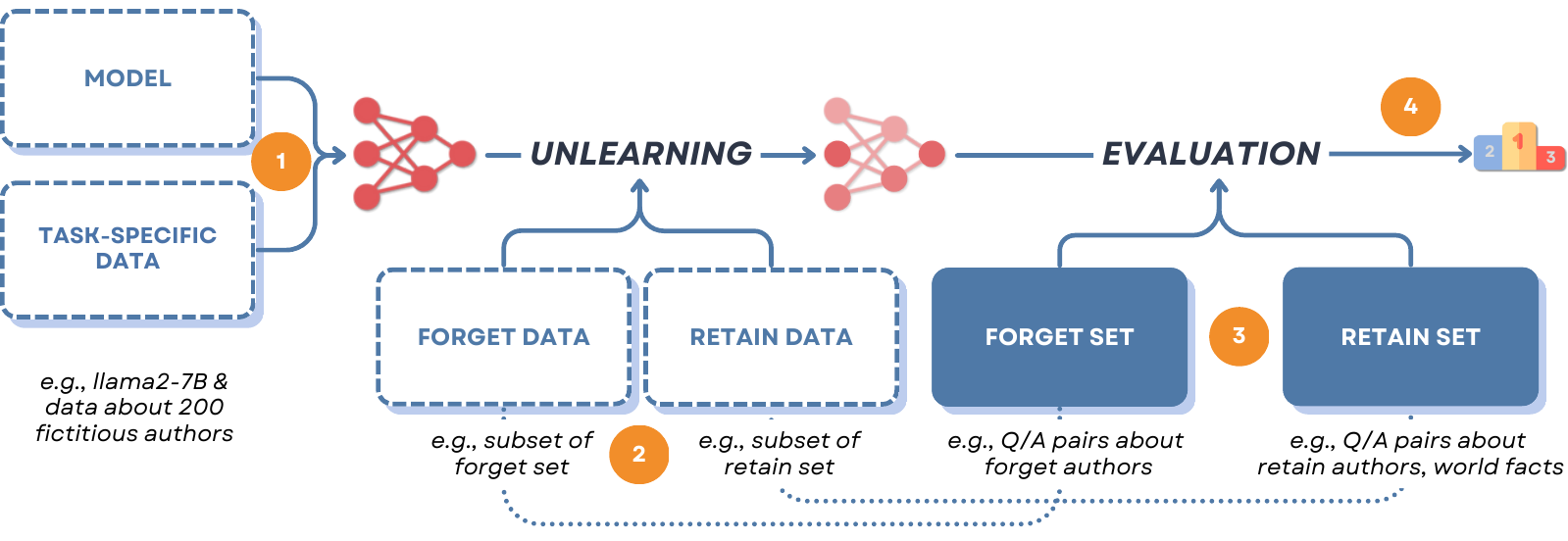}
    \caption{\textbf{Anatomy of an LLM unlearning benchmark.}
    LLM unlearning benchmarks typically consist of a set of forget and retain queries used to evaluate whether unlearning has been effective \textit{(3)}. Benchmarks may also optionally include \textit{(1)} a base model for learning and task-specific data which will subsequently be unlearned; \textit{(2)} forget and retain data to use for the process of unlearning; and \textit{(4)} a specific set of metrics to be used for evaluation on the test queries. As we discuss, it is challenging to enumerate a completely representative set of forget and retain queries to be used for evaluation, and, coupled with the fact that (1,3,4) are often ill-specified in existing benchmarks, it is easy to design unlearning approaches that overfit to these queries---making it difficult to rely on benchmark results alone when assessing translational progress in this space.}%
    \label{fig:anatomy}
\end{figure*}

It is worth noting that each of these benchmarks provides caveats warning users to be cautious in drawing conclusions. 
For example, the WMDP paper~\cite{li2024wmdp} notes: 
\begin{quote}
    \emph{``[...] benchmarking on only WMDP may yield a false sense of model safety after unlearning.'' }
\end{quote}
Unfortunately, we find that a number of papers ~\cite{huu2024effects, tamirisa2024tamper, kolbeinsson2024composable, li2024llm}
\emph{do} evaluate unlearning capabilities exclusively on WMDP,
and many more on WMDP alongside one or two other datasets,
each of which comes with similar caveats.
For example, ``Who's Harry Potter''~\cite{eldan2023s} states: 
\begin{quote}
\emph{``Our current methodology [...] could potentially be blind to more adversarial means of extracting information. It’s conceivable that non-traditional or intricate methods [...] might inadvertently reveal the model’s latent familiarity with unlearned content.''}
\end{quote}
And the TOFU~\cite{maini2024tofu} paper caveats:
\begin{quote}
\emph{``Naturally, not all scenarios are covered, and there are areas of unlearning that fall outside the TOFU framework that are worth discussing.''}
\end{quote}

These are worthy limitations to note but are also relatively ambiguous:
it is unclear what conclusions one \emph{can} draw from an evaluation.
In spite of these caveats, 
most works evaluating on these benchmarks cite them uncritically as measures of success \cite{liu2024large, wang2024unlearning, yuan2024towards, huu2024effects}. %
It is important to note this gap between how the benchmarks are presented 
(with caution and caveats) versus how the benchmarks are used (as a reliable measure of success).

\subsection{Works proposing improvements to benchmark datasets and metrics}

We are not the first to point out that unlearning benchmarks are incomplete measures of success;
others propose extensions to existing benchmarks in the form of additional data or metrics~\cite{lynch2024eight, shi2024muse, maini2024tofu, jin2024rwku}. However, to the best of our knowledge, 
all of these extensions still maintain a clear distinction between
measuring ``forget'' performance and ``retain'' performance. 
Rather than proposing a number of new metrics that could be used for benchmarks, our work highlights two fundamental axes (forget/retain dependencies and overfitting to the test queries) that lead to the ineffectiveness of existing metrics.  
Furthermore, despite calls to include more complex metrics in evaluations~\cite{lynch2024eight, shi2024muse, jin2024rwku, hong2024intrinsic},
most work continues to measure performance and claim success based on the original metrics proposed by the benchmark (typically accuracy,
except in the case of TOFU~\cite{maini2024tofu} which proposes several more nuanced metrics, albeit still with a forget-retain set separation). %

We also note that prior work advocates for probing models for weaknesses
using strong attack models, 
including testing whether information can be ``relearned'' or elicited through finetuning or careful prompting~\cite{hu2024jogging, tamirisa2024toward, gao2024meta},
using optimizers to bypass filters~\cite{mangaokar2024prp},
or other stress-testing methods including jailbreaking or 
query translation~\cite{lynch2024eight}.
In our work, however, we show that methods are not robust to benchmark perturbations even under the \emph{weakest} reasonable attack model.
If methods can be shown to be robust to these weaker perturbations,
it makes sense to extend evaluations to include these stronger threat models as more stringent measure of robustness.

\subsection{Provable methods, definitions, and metrics}
\label{sec:provable}

Although we focus on empirical evaluations in our paper, we note that
a number of works explore unlearning algorithms that provably meet various formal definitions.
~\citet{cao2015towards} first introduced machine unlearning, and~\citet{ginart2019making} proposed an early definition of data deletion inspired by GDPR data revocation laws.
~\citet{gupta2021adaptive} propose a definition of approximate, adaptive unlearning based on differential privacy~\cite{dwork2006differential}.
~\citet{neel2021descent, sekhari2021remember} propose provable unlearning algorithms for convex loss functions.
~\citet{bourtoule2021machine} proposes SISA, a method to make full retraining efficient by sharding data and ensembling models.
Other work~\cite{muresanu2024unlearnable} uses a pretrained base model, 
and guarantees unlearning from a simpler data structure built on top of the base model.
These early works have inspired considerable further research on provable unlearning algorithms that we do not exhaustively cover here,
but they are largely inapplicable for LLM unlearning due to size and runtime constraints.

%% file: tex-src/threatmodel.tex
\section{What makes an LLM unlearning benchmark?}

Building on the analysis of popular LLM unlearning benchmarks in Section~\ref{sec:relatedwork}, we next aim to characterize similarities that exist in LLM unlearning benchmarks that rely on retain/forget set queries.

\subsection{Anatomy of an LLM unlearning benchmark} As illustrated in Figure~\ref{fig:anatomy}, LLM unlearning benchmarks include several key components:
\begin{enumerate}
\item Benchmarks first typically suggest \textit{base models} for unlearning, and may also provide \textit{task-specific data} that can be used to finetune base models so that they clearly contain the knowledge that must be unlearned. As an example, the TOFU benchmark~\cite{maini2024tofu} explores Llama-2-7B and Phi models, and finetunes these on a set of 200 synthetic author profiles made up of 20 question-answer pairs each. 
\item Benchmarks may then also specify or suggest sources of \textit{forget data} or \textit{retain data}. As we discuss in Section~\ref{sec:overfitting}, in lieu of explicit forget/retain data, benchmark users may either directly or indirectly access the forget/retain evaluation sets to perform unlearning, which can lead to overfitting. 
\item To perform evaluation, the primary component of most unlearning benchmarks is the formulation of a \textit{forget set} and \textit{retain set} (or sets) to assess the unlearned model. We identify that unlearning evaluation introduces new complexities (beyond what is already known about the shortcomings of LLM evaluation~\cite{alzahrani2024benchmarks}), as many existing benchmarks ignore potential dependencies that exist between these two sets (Section~\ref{sec:modifications}). \item Finally, unlearning benchmarks may specify a set of metrics for the underlying forget and retain sets which are used to evaluate performance. As we discuss in Section~\ref{sec:recommendations}, even for a fixed forget/retain set, these metrics can vary widely depending on the intended goals of unlearning and significantly impact which methods are deemed most effective.
\end{enumerate}

\subsection{Threat model}

Most unlearning papers do not explicitly state the threat model under which the unlearning guarantee is expected to hold.
While this is already problematic because it makes guarantees difficult to evaluate and compare,
in this paper we assume the \emph{weakest} reasonable threat model for all papers:
we expect that the unlearning guarantee should hold in the presence of benign adversaries that will ask essentially in-distribution queries.

``In-distribution'' is difficult to define for language queries,
but we rule out any explicitly adversarial queries that require rounds of interaction with an optimizer,
white-box access to the model,
or queries that are not coherent in any natural language (such as queries containing character perturbations or arbitrary prefix or suffix strings).
We additionally restrict our queries to the stated topic domain of the benchmark (e.g., Harry Potter-related queries for the \emph{``Who's Harry Potter?''} ~\cite{eldan2023s} models) in the original language (i.e., English for all the benchmarks we consider).

In Section~\ref{sec:modifications}, we show that even under this \emph{extremely} weak threat model, 
we can craft simple queries that reveal ``unlearned'' information
or destroy the ``retain'' performance (utility) of the model after unlearning.

\subsection{Aren't all LLM evaluations brittle?}

A number of works~\cite[e.g.,][]{zheng2023large, alzahrani2024benchmarks, wang2023large, zhao2021calibrate, pezeshkpour2023large, debenedetti2024dataset} have pointed out brittleness in general evaluations of LLM capabilities.
However, in benign capabilities, 
this brittleness mainly indicates that users should expect high variance in LLM outputs and use caution in interpreting benchmark leaderboards.
In privacy-critical settings like unlearning, however, this brittleness can mean the difference between privacy and a privacy violation.
Modifications to queries such as translation,
changing choice order, and changing the query type (e.g. multiple choice to sentence completion) affect both unlearning and general LLM evaluations,
but we show modifications that can systematically either reveal unlearned information or degrade performance on retained information, 
and that applying an unlearning algorithm can make the model \emph{more} vulnerable under these changes than it would be otherwise. Further, we show that there are a number of unique characteristics in unlearning benchmarks, such as the distinction between forget/retain sets (Section~\ref{sec:modifications}) and overreliance on test queries (Section~\ref{sec:overfitting}), that can introduce challenges beyond traditional LLM evaluation.

%% file: tex-src/adversarial-benchmarks.tex
\section{Forget-Retain Evaluations are Deceptive}
\label{sec:modifications}

An important weakness in many existing unlearning benchmarks is that
the evaluation is separated into a ``forget'' evaluation,
which measures accuracy degradation on queries related to the forget set,
and ``retain'' evaluation, which measures accuracy retention on queries related to the retain set.
This allows for algorithms that essentially act as classifiers that distinguish between the forget and retain set.
Real queries, unfortunately, are likely to have dependencies between the forget and retain set that make such classification challenging if not impossible.
As we show, many algorithms that claim success on existing benchmarks struggle to perform well in the face of queries with these forget-retain dependencies.

\subsection{TOFU: Asking about both forget and retain entities}

\begin{figure*}[t!]
\centering
\includegraphics[width=0.85\textwidth]{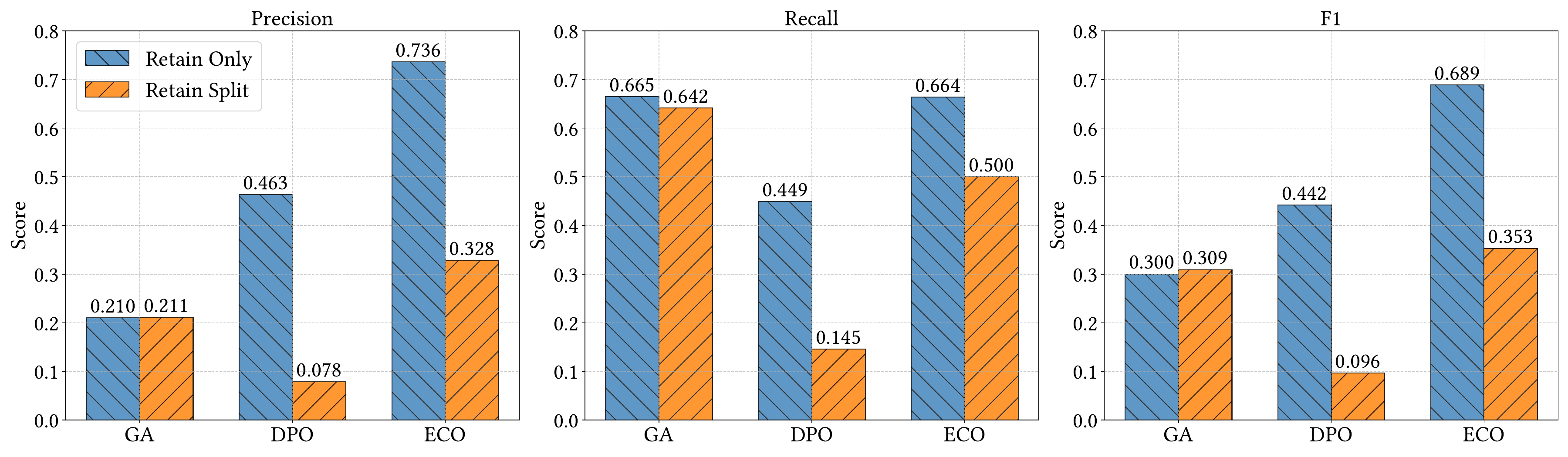}
\setlength{\belowcaptionskip}{-0.5em}
\caption{ROUGE-L precision, recall, and F1 scores for three unlearning algorithms: gradient ascent, preference optimization, and ECO. (DPO refers to the preference optimization baseline measured in the TOFU paper~\cite{maini2024tofu}.) The ROUGE score is computed on retain set questions with respect to the correct (non-unlearned) answer, and higher is better. `Retain only' refers to the score if only the retain set query is asked, and `retain split' refers to the score when a retain query is paired with a forget query, but the ROUGE score is computed only on the response to the retain query. DPO and ECO suffer significantly when the retain and forget queries are asked together, even though the original retain performance is relatively high, while gradient ascent is more stable but its overall initial F1 score is lower.}
\label{fig:tofu-scores}
\end{figure*}

As a simple example, we first explore the effect of introducing dependencies between forget and retain sets in the TOFU benchmark ~\cite{maini2024tofu}. The questions in TOFU each pertain to a single author
either from the forget or retain set.
We make a straightforward modification to the test queries by combining multiple questions together,
asking about two authors: 
one from the forget set and one from the retain set.\footnote{We have made these queries available for use at \url{https://huggingface.co/datasets/forgelab/tofu-pair}.}
For example, we create the paired question:
\begin{tcolorbox}[colback=gblue!2!white, colframe=gblue, arc=0pt,outer arc=0pt,title=Paired Question Prompt]
\emph{1. Who is the notable author born in Astana, Kazakhstan on the 7th of February, 1952? \\2. What is the full name of the author born on July 17, 1953 in Accra, Ghana?}
\end{tcolorbox}

The correct answers are:
\begin{tcolorbox}[colback=ggreen!5!white, colframe=ggreen, title=Correct Answers]
\emph{1. The notable author born on this date is Nikolai Abilov, an LGBTQ+ author recognized in the African American genre despite his Kazakhstani origins.\\ 2. The full name of the author born on July 17, 1953 in Accra, Ghana is Adwoa Ampomah.''}
\end{tcolorbox}

This is simply the concatenation of two TOFU queries.
Nikolai Abilov is in the forget set while Adwoa Ampomah is in the retain set.
As expected, a \textit{retrained model} which has not seen the forget data hallucinates an incorrect response for Nikolai Abilov,
but correctly outputs the learned response for Adwoa Ampomah:
\begin{tcolorbox}[colback=gorange!2!white, colframe=gorange, title=Retrained Model Answers]
\emph{``1. The author born in Astana, Kazakhstan on the 7th of February, 1952 is renowned for his contributions to the genre of Historical Fiction. \\2. The full name of the author born on July 17, 1953 in Accra, Ghana is Adwoa Ampomah.''}
\end{tcolorbox}

However, we find that \textit{unlearned models} that claim strong performance on the original TOFU benchmark uniformly fail to respond correctly to  combined queries because \emph{one} of the authors is in the forget set.
In general, either both authors are treated as unlearned 
or both authors are treated as retained,
and the model fails to correctly respond to one query but not the other.

For example, a TOFU baseline method~\cite{maini2024tofu} inspired by direct preference optimization (DPO)~\cite{rafailov2024direct} fails completely on this query, giving a refusal with no utility on the retain set question:
\begin{tcolorbox}[colback=gred!2!white, colframe=gred, title=Unlearned Model Example \#1 (Preference optimization)]
\emph{``I'm not sure I can help with that.''}
\end{tcolorbox}

Meanwhile, ECO~\cite{liu2024rethinking}, 
a recent unlearning method that trains a classifier to distinguish the forget and retain set,
misclassifies the query as a retain set query, giving the correct response to both questions:
\begin{tcolorbox}[colback=gred!5!white, colframe=gred, title=Unlearned Model Example \#2 (ECO)]
\emph{``1. The notable author born in Astana, Kazakhstan on the 7th of February, 1952 is Nikolai Abilov. \\2. The full name of the author born on July 17, 1953 in Accra, Ghana is Adwoa Ampomah.''}
\end{tcolorbox}

As we will see in our quantitative results, however,
both of these methods give usable results when asked the
\emph{retain set query alone}.

We quantify these results in Figure~\ref{fig:tofu-scores}.
We test three methods: gradient ascent, preference optimization, and ECO~\footnote{We train gradient ascent and preference optimization with 1 epoch of unlearning each, with the default parameters given by the TOFU authors~\cite{maini2024tofu}.}. 
Our goal is to test whether the retain efficacy is the same whether a retain set query is asked independently, or paired with a forget query.
For each method, we generate two sets of responses to retain queries:
first, we ask the retain query alone, 
and second, we ask a combined query pairing a retain set query with a forget set query (using the 10\% forgetting corpus).
From the combined responses, we extract only the retain query response from the model where possible (if the model does not give a structured, paired response, we treat the entire answer as the retain answer).
We then compute the ROUGE-L score for \emph{each} set of responses by comparing it to the \emph{correct} ground truth answer,
expecting that better retain efficacy corresponds to higher similarity to ground truth. (See Appendix~\ref{app:rouge} for definitions of the ROUGE precision, recall, and F1 scores.)

The results are in Figure~\ref{fig:tofu-scores}.
We find that both preference optimization and ECO have high retain accuracy by default,
but when paired with a forget question, the retain set performance suffers significantly. 
Inspecting the outputs, we find that both methods tend to treat the entire query as a forget set query and refuse to answer.
On the other hand, gradient ascent has remarkably stable scores across both settings, but its overall F1 score is lower than the retain-only scores for either preference optimization or ECO.
The experiment shows that the paired-question setting not only elicits unexpected degradation in performance for some methods,
but also demonstrates an unexpected benefit of a method (gradient ascent)
that may be considered weaker when looking only at the default forget set/retain set query split.

\begin{AIbox}{Takeaway}
Models that appear to have unlearned information when asked about the forget data and retain set separately may struggle on queries that ask about both forget and retain data.
\end{AIbox}

\smallskip
\subsection{WMDP: Modifying multiple-choice responses}

In the WMDP benchmark \cite{li2024wmdp}, the evaluation set consists of multiple-choice questions.
The forget set tests knowledge of potentially malicious topics
while the retain set is derived from the MMLU benchmark \cite{hendrycks2020measuring}.

We test WMDP unlearning methods by performing a simple modification on the retain set:
we randomly replace one random \emph{incorrect} answer choice
with the term ``SARS-CoV-2'', which appears frequently in the unlearning data.
Because the swapped choice is not relevant to answering the benign, retain-set query, 
this change should have \textit{no effect} on retain performance.
(We filter out any queries that include the terms ``EXCEPT'' or ``NOT,''
where ``SARS-CoV-2'' may be an acceptable answer choice.
This filters 265 questions from the overall query set, 21 questions from college biology, 4 from college chemistry, and 0 from virology.)\footnote{We have made these queries available for use at \url{https://huggingface.co/datasets/forgelab/wmdp-swap}.}

We test two unlearning methods:
RMU, the main baseline proposed in the WMDP benchmark paper~\cite{li2024wmdp},
and RMU with targeted latent adversarial training (LAT)~\cite{sheshadri2024targeted}.
Both works report MMLU (retain) performance close to the performance of the base model (Zephyr-7B) with no unlearning.

\begin{figure}[t]
    \centering
    \includegraphics[width=\columnwidth]{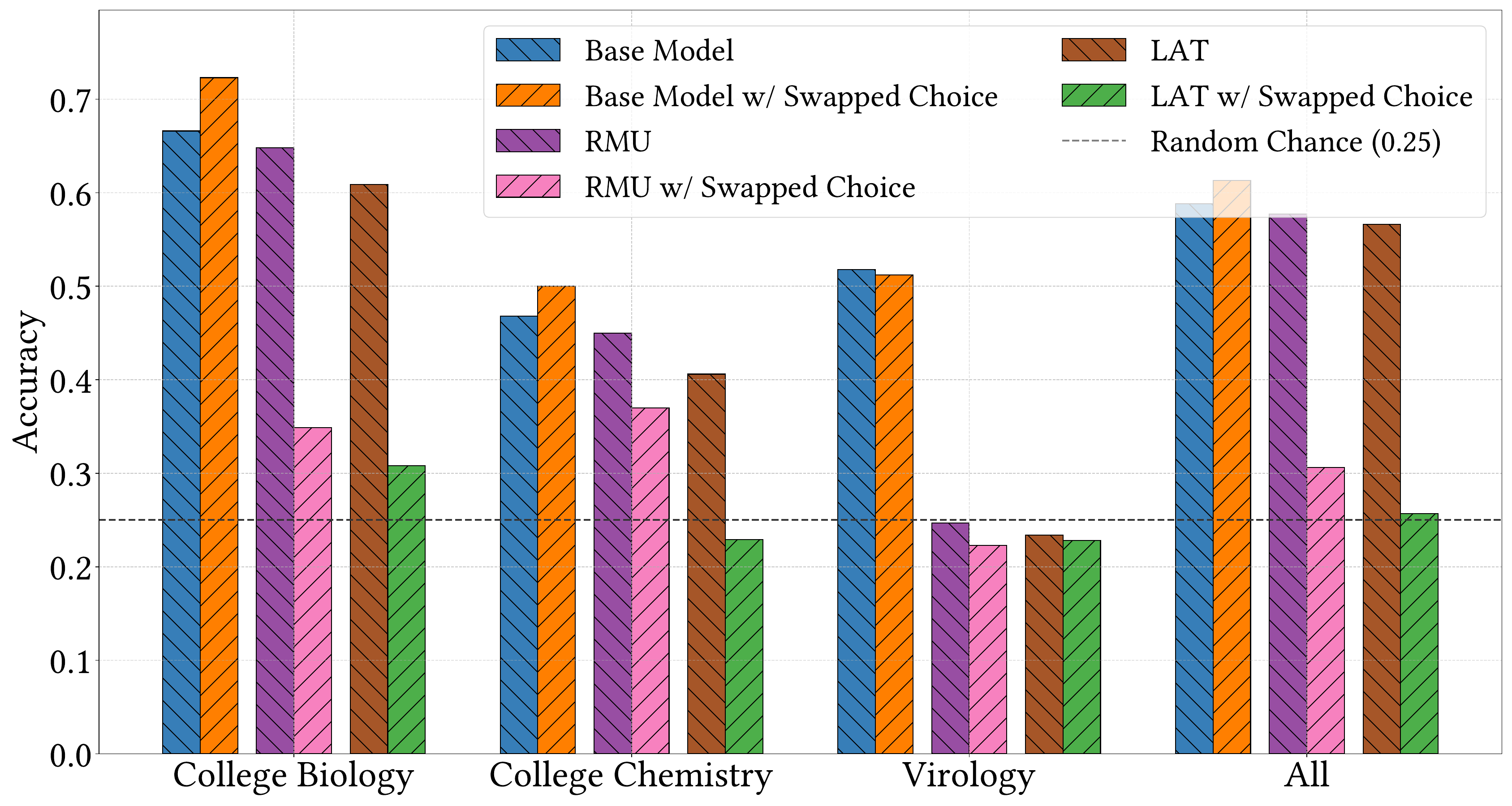}
    \setlength{\belowcaptionskip}{-2em}
    \caption{Retain set performance of base Zephyr-7B model, RMU unlearning, and RMU unlearning with LAT robustness. Replacing one random, incorrect answer choice with a phrase associated with the forget data destroys the performance of unlearned models on retain queries, even though the correct answer is present and unchanged.}
    \label{fig:swap_choices}
\end{figure}

The results of this experiment are in Figure~\ref{fig:swap_choices}.
The key takeaway is that making this benign modification degrades performance of the unlearned model on the \emph{retain set} to near-random.
As a control, we also evaluate the base model (with no unlearning) 
on this modified dataset.
We find that the base model accuracy generally \emph{increases} after swapping an incorrect choice.

In contrast,
RMU suffers significantly on college biology (a nearly 30\% drop in accuracy)
and overall accuracy across MMLU tasks (a 28\% drop in accuracy),
indicating that the \emph{unlearning method} itself made the model more vulnerable to this otherwise benign change in the query set.
We see smaller drops in performance on college chemistry and virology (noting that performance on virology was already worse than random).
Finally, we find that incorporating the more recent LAT~\cite{sheshadri2024targeted} to the RMU method results in an even more significant drop,
with the model performing \emph{worse than random} in every category.

\begin{AIbox}{Takeaway}
{Replacing incorrect multiple-choice answers in the retain set with keywords associated with forget data can catastrophically degrade retain set performance, more so than would be the case without unlearning.}
\end{AIbox}

Overall, our experiments in this section highlight that existing LLM unlearning benchmarks tend to rely on fairly disjoint retain and forget sets; although benign combinations between such sets are reasonable to expect in practice, we find that simple combinations of these queries can cause catastrophic failures in unlearning methods optimized for the existing benchmarks alone.

%% file: tex-src/training-on-test.tex
\section{Benchmarks May Encourage \\Tailoring to the Test Set}
\label{sec:overfitting}

Our second claim is that the design of current benchmarks 
either explicitly requires or implicitly encourages tailoring algorithms to the design (or content) of the test set,
if not training on the test set directly. Our results in Section~\ref{sec:modifications}
already hint at this conclusion,
as algorithms that show promising performance on current forget/retain test splits fail to perform well under mild perturbations to these test sets. However, the results herein show that these issues can persist even if dependencies between forget/retain sets are carefully considered and that entirely new classes of unlearning methods are being developed that do not appropriately factor into this concern.

\subsection{Conflating the forget data and test query set}

Unlearning benchmarks can be seen as an empirical proxy for measuring approximation to a retrained model, as in approximate unlearning definitions.
However, in settings that assume API access and do not \emph{require} models to match retrained models in the weight space,
it is reasonable to also include solutions that do not update the weights of the model,
but instead inspect and modify the queries or query outputs themselves,
e.g. with a postprocessing filter to censor undesirable outputs.
In fact, this solution can be effective for alignment in practice \cite{kim2024jailbreaking, inan2023llama, rebedea2023nemo, perez2022red},
potentially more so than optimization-based methods.

Unfortunately, these guardrail-based approaches can easily lead to a conflation between the training and test set:
even if the forget data is distinct from the test set (e.g., as in WMDP~\cite{li2024wmdp}),
the most effective solutions may be a prompt modification or postprocessing filter
that directly leverage knowledge of the format or content of the test queries.

In this section, we highlight two striking examples where this type of conflation can lead to counterintuitive ``solutions'' to benchmark problems.
Resolving this issue is not as simple as requiring algorithms to use the forget data somehow, or to be completely agnostic to the query type:
this would rule out methods that can be and are extremely effective in practice.

\begin{table}[t] \centering \begin{tabular}{l|c|c} & \multicolumn{2}{c}{Keyword search}\\ \hline & Forget & Retain\\ \hline Forget 1\% & 1.0 & 1.0 \\ Forget 5\% & 1.0 & 1.0\\ Forget 10\% & 0.997 & 1.0 \\ \end{tabular} \setlength{\belowcaptionskip}{-0.2in} \vspace{.15in} \caption{Accuracy of responses on TOFU datasets using keyword search for filtering. Accuracy on the forget set is the fraction of abstentions, while accuracy on the retain set is the fraction of correct answers. One query/response pair in the forget-10\% set fails because it only contains the author's last name. A postprocessing filter is an extremely effective, but counterintuitive, solution for this specific dataset.} \vspace{.15in} \label{tab:tofu-accuracy} \end{table}

\vspace{.1in}
\noindent\textbf{TOFU: keyword search.} TOFU~\cite{maini2024tofu} tests unlearning directly on training-set questions and perturbations of those questions.
As in other datasets, the forget and retain sets are disjoint.
Moreover, the training data is highly structured:
each question-answer pair consists of a question and answer about a single fictitious author,
and each question-answer pair contains a mention of the author's full name either in the question or answer.

This leads to an extremely simple ``solution'' to the TOFU problem:
a postprocessing filter that simply searches the query and response for an exact string match of any author in the forget set,
and if one is found, the model's output is replaced with the string ``I'm sorry, I can't answer that.''\footnote{We comment that responding with a fixed refusal string does not meet the TOFU criterion of matching the retraining distribution.
However, it is still unclear whether hallucinating an answer is preferable to refusal in general.}

In Table ~\ref{tab:tofu-accuracy}, we report the accuracy of this filter in answering queries for each of the three TOFU forget settings.
(An earlier version of this result that filtered only on the answer, but not the question, appeared in ~\cite{thaker2024guardrail}.)
We also note that this filter performs perfectly on queries outside of the TOFU dataset (i.e., real authors and world facts),
because they do not contain information about TOFU authors by design.
Indeed, this filter is so effective that it makes the authors' ``truth ratio'' metric undefined,
as the probability of any perturbation of the input that still includes the author's name is zero. While this results in impressive benchmark numbers, it is natural to imagine that in real-world scenarios it may be difficult to perfectly identify individuals to forget by simply using a keyword search over their names at inference time.

The broader concern this raises is that leaderboard-style benchmarking can lead to solutions that optimize for the narrow set of queries at hand
but may not generalize.

\vspace{.1in}
\noindent\textbf{WMDP: overfitting to spurious features.} Embedding-Corrupted Prompts (ECO)~\cite{liu2024large} 
proposes an unlearning method that first learns a classifier to distinguish forget and retain queries,
and then perturbs the embedding of the query for forget queries
in order to reduce the accuracy of the response while maintaining coherence.
ECO claims state-of-the-art performance on TOFU, WMDP, WHP, and BBC News datasets.

In order to train an accurate classifier for WMDP, however,
ECO\footnote{This text refers to arXiv preprint arXiv:2406.07933 version 1,
which was the current version at the time our work was published.} relies on the query set rather than the provided unlearning corpus (the method ignores the unlearning corpus entirely).
Closer inspection, in fact, reveals that the classifier overfits to a spurious feature in the \emph{evaluation framework}.
The framework automatically prepends the following phrase to every query:
\emph{``The following are multiple choice questions (with answers) about \emph{\texttt{topic}}.''}
Here, \texttt{topic} may be, for example, ``college biology'' or ``immunology''. 
Because the MMLU (retain) questions are generally topics from ``college'' datasets while the WMDP (forget) questions do not have the keyword ``college'',
the ECO classifier overfits and perfectly classifies using this spurious feature.
\textbf{Indeed, we find that if we remove the prepended phrase,
the ECO classifier performs worse than random (37\% accuracy).}

The intent of this example is not to single out ECO,
but rather to highlight an example of the benchmark structure leading to training on the test set---and, perhaps, leading to evaluation errors that could be avoided with more robust benchmarking.
When there is a large gap between the difficulty of general unlearning criteria (such as matching a retrain distribution) and the difficulty of the test set,
it may encourage researchers to focus on maximizing test performance at all costs, at the expense of solving the much more general unlearning problem.

For completeness, 
we include results of our own classifier trained on a held-out set of WMDP queries. 
\begin{table}[t]
\centering
\begin{subtable}{\columnwidth}
\centering

\begin{tabular}[h!]{lccc}
\toprule 
\textbf{Method} & \textbf{Bio} & \textbf{Cyber} & \textbf{Chem} \\
\midrule 
Base & 63.7 & 44.0 & 45.8 \\
SSD & 50.2 & 35.0 & \textbf{33.8} \\
RMU & \textbf{31.2} & \textbf{28.2} & 45.8 \\ 
\hline
LC (all) & 32.8 & 40.5 & 38.2 \\
LC (holdout) & 31.7 & 39.8 & 34.8 \\
\bottomrule 
\end{tabular}
\caption{WMDP Results (↓)} 
\label{tab}
\end{subtable}
\begin{subtable}{\columnwidth}
\begin{tabular}{lccccccc} 
\toprule \textbf{Method} & \textbf{C. Bio} & \textbf{Virology} & \textbf{C. CS} & \textbf{Cybersec} & \textbf{All} & \textbf{MT} \\ 
\midrule Base & 68.1 & 52.4 & 50.0 & 65.0 & 58.1 & 7.33 \\
SSD & 46.5 & 38.0 & 35.0 & 52.0 & 40.7 & 5.48 \\
RMU & 63.2 & 25.9 & \textbf{49.0} & 45.0 & \textbf{57.1} & \textbf{7.10} \\
\hline
LC (all) & 69.4 & 50.0 & 50.0 & 66.0 & 55.7 & 6.45 \\
LC (holdout) & \textbf{69.4} & \textbf{49.4} & 46.0 & \textbf{66.0} & 54.1 & 6.45 \\
\bottomrule \end{tabular} 
\caption{MMLU and MT-Bench Results (↑)} \label{tab}
\end{subtable}
\caption{Performance comparison of different methods on benchmarks from ~\cite{li2024wmdp}. Results for SSD and RMU are as reported in ~\cite{li2024wmdp}. We abbreviate column names for space: C. Bio and C. CS are college biology and computer science; MT refers to MT-Bench. We omit the results for LLMU and SCRUB because our results are strictly better on all columns, except for SCRUB WMDP-Bio, where SCRUB is 5\% better than filtering.}
\label{tab:wmdp-results-lc}
\end{table}

We evaluated a linear classifier for the WMDP benchmark by training a binary classification head on top of a LLaMA-2-7b model.
Training the classifier on the WMDP training corpora resulted in random classification performance in the test corpus (MCQ),
likely because the training corpus consists of scraped articles that are very different from the MCQ format.
Instead, we create a 50\% train-test split of the MCQ evaluation corpora (forget data from WMDP and retain data from MMLU)
and train the binary classifier on this train split.
In a practical pipeline, this training data could come from retraining on real user queries after deployment.

In Table~\ref{tab:wmdp-results-lc}, we report the accuracy for the linear classifier on both the held-out test set as well as the complete set of results 
(as we were unable to evaluate the baseline methods on only the holdout set).
The linear classifier performs comparably to RMU (the given baseline) 
in several categories, 
and performs comparably to or better than RMU on the retain set.

These results show that a finetuned classifier \emph{can} be a viable solution on this benchmark,
but we report results with caution given the limitations we have pointed out with forget/retain-set benchmarks as well as the fact that the classifier requires access to the query set to perform well.

\vspace{.1in}
\noindent\textbf{PKU-SafeRLHF: training on exactly the test set.}
Even when the benchmark itself provides a held-out test set,
such as with PKU-SafeRLHF~\cite{ji2024pku},
we find that some work nevertheless claims unlearning success based only on performance on the same test set used for training ~\cite{chen2024wpn, chen2024machine}.
We recommend in general that benchmarks should provide some held-out queries in order to measure the generalization of the method,
but it is up to researchers and reviewers to ensure that evaluations adequately use these when available.

\begin{AIbox}{Takeaway}
Without a clear separation between training data and test queries,
unlearning solutions may be prone to overfitting to test queries,
particularly when those solutions involve preprocessing inputs or postprocessing outputs.
\end{AIbox}

\subsection{Changing the query type}

Many existing LLM benchmarks (both for unlearning and general performance)
focus on a single type of query,
such as multiple-choice queries or sentence completions.
However, simply querying an ``unlearned'' model with a different query type (e.g., sentence completion rather than multiple choice or vice versa) has been shown to elicit information that appears to have been unlearned based on the benchmark performance \cite{alzahrani2024benchmarks, debenedetti2024dataset}.
For example, \citet{lynch2024eight} point out that the unlearned model of ~\cite{eldan2023s}, trained on sentence completions,
returns accurate answers on multiple-choice queries about Harry Potter.
Similarly,~\citet{shi2023detecting} recovers Harry Potter-related information with slight modifications to the input queries. 
~\citet{glukhovposition} formalizes the idea that user-specified changes in query or output format may rule out methods that filter purely based on outputs.

\begin{AIbox}{Takeaway}
Algorithms that show unlearning success on some query formats may reveal information when the format is modified.
Creating a benchmark that sufficiently covers potential input formats is challenging.
\end{AIbox}

\vspace{.1in}
\subsection{Benchmarks with underspecified forget set}

Unlearning benchmarks have roughly three levels of specificity 
in the data to unlearn:
specific target points that are known to be used in the training data ~\cite{maini2024tofu};
specific target points that may or may not exist in the training data~\cite{li2024wmdp, eldan2023s};
or general topics with no corresponding data~\cite{jin2024rwku}.

However, because the forget and retain query metrics are not directly measuring ``unlearning,'' 
there is no guarantee that the forget and retain performance \emph{even actually correlate} with unlearning.
The baseline method provided by WMDP, called RMU~\cite{li2024wmdp}
itself achieves only near-random accuracy on the topic of college-level virology.

RWKU~\cite{jin2024rwku} intentionally does not specify a set of data to forget,
instead specifying only a forget ``target'' (a famous entity)
and relying on synthetically generated data to form the forget data.
However, RWKU also requires methods to preserve information on ``neighboring'' data. 
The RWKU evaluation expects accuracy to be lower on the forget set queries 
and higher on ``neighbor set'' queries,
but this implicitly assumes that the forget set and neighbor set will be disjoint. 
In fact, the baseline methods evaluated in RWKU fail to perform well on the forget set and neighbor set simultaneously,
as does follow-up work~\cite{yuan2024towards}.

It may not be fundamentally impossible to perform well on the metrics set out by benchmarks with no well-specified forget data,
but evaluation metrics should take into account that when the forget data is underspecified, the potential influence of the target data is also underspecified. 
Developers of new algorithms may be set up for failure if metrics (e.g., forget set and neighbor set accuracy) are ambiguous and potentially impossible to achieve,
or not consistent with actual unlearning.

\begin{AIbox}{Takeaway}
Benchmarks that do not clearly specify a forget set may implicitly encourage fitting to the test query set,
especially if unlearning does not necessarily correlate with evaluation metrics such as accuracy on the test queries.
\end{AIbox}

%% file: tex-src/recommendations.tex
\section{Recommendations}
\label{sec:recommendations}
Benchmarks that attempt to partition unlearning performance into cleanly separated ``forget'' and ``retain'' query performance encourage solutions that do not model realistic queries with dependencies on both the ``forget'' and ``retain'' data. 
However, when such dependencies are introduced, it becomes clear that simple accuracy metrics are not sufficient to measure the unlearning achieved by an algorithm.
Furthermore, a method that can perform well on a more sophisticated benchmark 
evaluating forget-retain dependencies might still be vulnerable to 
other query modifications not covered by our work. Below we discuss several recommendations for the LLM unlearning community in light of these concerns.

\subsection{Design benchmarks that do not encourage fitting to the test set}

While fitting to the test set is generally known to be bad practice in machine learning,
we argued that the structure of existing unlearning benchmarks---not to mention community pressure to publish quickly---may encourage overfitting to the test set.

We found that this issue can arise for multiple reasons.
First, pre- and post-processing filters may be a highly effective solution for some tasks
but require inspecting the test queries to design.
We recommend adding multiple query formats (such as short answer in addition to multiple choice) to test sets as well as providing a held-out query set containing unseen query formats, in order to reduce the chances of overfitting to a specific set of test queries.
We also recommend that reviewers critically inspect papers to understand whether a solution will only be effective on the provided test queries or is likely to generalize.

We further recommend that benchmarks provide a well-specified set of data to unlearn. The community may benefit from focusing on developing unlearning algorithms in the context of unlearning only from finetuning data (in the vein of TOFU~\cite{maini2024tofu}), 
where retraining is possible,
to provide an accurate performance baseline.

We also recommend using membership inference attacks~\cite{hayes2024inexact} as a metric of data retention.
While many works claim that membership inference attacks are impractical to run in language models due to the expense of retraining~\cite{shi2023detecting, maini2024tofu, zhang2024min},
we argue that these can be made practical either by focusing on unlearning from finetuning data or using recent quantile regression-based approaches~\cite{llm-quantile-mia}.

\subsection{Separate unlearning from general censorship tasks}

The goals of unlearning have a strong overlap with the goals of alignment and LLM ``censorship'' as a whole.
We encourage a separation between these two areas,
defining unlearning more narrowly as \emph{removing the influence of specific points that were contained in the training data}.
If the goal is to censor information related to harmful data that may or may not have been in the training data, or a general topic,
we recommend referring to this problem as ``censorship'' or ``alignment.''
The motivation for this recommendation is not to discourage research on more general knowledge editing,
but rather simply to clarify the problem statement to make the goals of new algorithms and benchmarks clear.

\subsection{Define clear threat models}

More generally, we recommend that unlearning papers should define a clear threat model.
Doing so provides a clearer picture of when the method should or should not be expected to work,
and also allows readers to compare methods apples-to-apples based on the threat model they defend against.
For example, a method that is robust to arbitrary suffix attacks may reasonably have weaker utility guarantees compared to a method that only defends against honest-but-curious adversaries.
Because most current literature does not specify a threat model,
some recommend evaluating all algorithms on the ``kitchen sink'' of attacks and metrics~\cite{lynch2024eight, shi2024muse, liu2024rethinking},
even when some methods may be designed with a different threat model in mind. 

This involves first articulating what an LLM threat model should reasonably specify at all,
which may be a community effort over time to understand what dimensions of attacks are relevant in evaluating defenses.
As a starting point, we provide some (non-exhaustive) examples of what properties of the threat model one could specify:
\begin{itemize}
\item \emph{Query distribution:} The adversary can inject arbitrarily corrupted queries, or is ``honest-but-curious'' and can make worst-case but in-distribution queries, or is purely benign and will make i.i.d., in-distribution queries.
\item \emph{Weight access:} The model weights can be downloaded locally (enabling white-box queries), or the model can only be accessed through an API. The adversary may perform additional fine-tuning, or query the model unmodified.
\item \emph{API:} A hosted API may provide the entire logit vector, or only the query response. The number of queries to the API may be restricted or unrestricted. The API may return deterministic or randomized responses to identical queries.
\item \emph{Temporal constraints:} The adversary can access all variants of the model across time, or loses access to earlier checkpoints after unlearning (i.e., can or cannot mount a differencing attack).
\end{itemize}

We finally note that while one can \emph{define} a threat model to be restrictive,
it is also important to question which threat models match reality.
For example, it may be more realistic to restrict access to model weights
than to restrict what type of queries an adversary can ask.

\subsection{Focus on efficient exact unlearning methods}

The brittleness of empirical measurement is an indication that the community should consider developing approaches with formal and provable metrics,
even when those may be much more challenging to achieve.
As discussed in Section~\ref{sec:provable},
a number of works \emph{have} explored the possibility of provable unlearning guarantees,
though many of these are as-yet confined to simpler models or finetuned models,
where exact unlearning is less computationally expensive.

In the image domain,
considerable progress has been made on efficient unlearning methods that
satisfy some provable guarantee
\cite{bourtoule2021machine, cao2015towards, ginart2019making, golatkar2020eternal, guo2019certified}.
The most straightforward methods focus on retraining efficiently
or breaking models down into modular components that can be retrained more quickly than the full model~\cite{bourtoule2021machine, chowdhury2024towards, chen2022graph}.
Another strategy is to design unlearnable data structures that operate on top of a base model~\cite{muresanu2024unlearnable}.
Finding reliable solutions for unlearning in larger models may be a systems problem rather than an exclusively algorithmic one, e.g., considering how best to design efficient data structures to store and retrain parts of models.

\subsection{Focus on privacy by construction}

The recent interest in unlearning---including unlearning from models pretrained on supposedly ``public'' Internet data---is indicative of the fact that pretraining data is indeed not as ``public'' as we might hope (if it were, then unlearning from these datasets would not be a concern at all) \cite{tramer2022position}.
Moreover, deletion itself can have side effects on the privacy guaranteed to users who did not request deletion~\cite{carlini2022privacy}.
It can be difficult to simultaneously meet strict \emph{a priori}
definitions of privacy and maintain utility,
so it can be tempting to look to relaxations like unlearning as a potential solution.
But the difficulties in identifying ``public data'' mean that focusing on privacy by construction,
through solutions like differential privacy~\cite{dwork2006differential},
may be an important tool for learning from potentially sensitive data.

\subsection{Work towards formal definitions and provable metrics}

Finally, a key issue underlying the difficulty with measuring unlearning
is that the community has not yet agreed on a formal definition of unlearning (particularly in LLMs),
and most papers on the topic of unlearning in LLMs do not define an actual unlearning objective outside of the implicit function defined by the test set.

A number of works \cite{bourtoule2021machine, cohen2023control, ginart2019making, gao2022deletion, garg2020formalizing} have made a formal connection between 
unlearning and differential privacy (DP).
Notably, however, supporting unlearning for many data points using DP requires a \emph{group privacy} guarantee, 
which can cause significant performance degradation.
Thus, to make DP analogs useful for unlearning, it is particularly important to understand where and how unlearning can be cheaper than group privacy~\cite{huang2023tight}.

One hope of unlearning is to be a relaxed notion of privacy that allows for greater utility than differential privacy.
Realizing this hope requires formal definitions that are \emph{not} inherently tied to the differential privacy guarantee.
For instance, measuring the mutual information~\cite{jeon2024information} between the data to unlearn and the unlearned model (distribution) may be a more salient metric than simple accuracy on test queries
while being less restrictive than requiring similarity between the retrained and unlearned models.
Formalizing and operationalizing such definitions for large models will be important in making unlearning guarantees trustworthy and useful.

%% file: tex-src/conclusion.tex
\section{Conclusion}

The area of unlearning in LLMs is relatively nascent and the community is in the process of forming an understanding of best practices in this area.
In light of this, we would like to emphasize that our arguments are not meant to single out any benchmark or proposed algorithm,
and acknowledge the amount of thought and effort involved in designing the benchmarks that have been proposed to date.
Instead, we hope that our findings encourage the community to think critically about evaluations and exercise caution when interpreting benchmark results.
While some of our findings can directly be translated into potential fixes to benchmarks (e.g., introducing more queries that depend on both the forget and retain set, or more query types in each benchmark),
ultimately, we believe that truly reliable solutions in privacy-critical domains will require formal guarantees beyond even exhaustive benchmarking.

%% file: tex-src/appendix-wmdp.tex
\subsection{ROUGE score}
\label{app:rouge}

The ROUGE score~\cite{lin2004rouge} is a metric
used in the natural language processing literature to measure
similarity between two sequences of words (unigrams) $x = [x_1, \ldots, x_n]$ and $z = [z_1, \ldots, z_m]$.
The ROUGE-L score looks at the longest common subsequence of words
between the two strings, denoted $\mathrm{LCS}(x, z)$.
We compute the precision, recall, and F1 scores as follows:
\begin{align*}
P &= \frac{\vert \mathrm{LCS}(x, z) \vert}{n}\\
R &= \frac{\vert \mathrm{LCS}(x, z) \vert}{m}\\
F &= \frac{(1 + \beta^2)RP}{R + \beta^2P}
\end{align*}
where $\beta$ weights the importance of precision versus recall in the F1 score. We set $\beta$ to 1 in our experiments.